\documentclass[runningheads]{llncs}
\usepackage[T1]{fontenc}
\usepackage{graphicx}

\usepackage{cite}
\usepackage[english]{babel}
\usepackage{amsmath,amssymb,amsfonts}
\usepackage{algorithmic}
\usepackage{graphicx}
\usepackage{textcomp}
\usepackage{multicol}
\usepackage{multirow}
\usepackage{comment}
\usepackage{xcolor}
\usepackage{url}
\usepackage[normalem]{ulem}

\begin{document}

\title{Predictive Process Monitoring Using Object-centric Graph Embeddings\\
}
\author{Wissam Gherissi\inst{1}\orcidID{0000-0002-2234-2963} \and Mehdi Acheli \inst{2} \orcidID{0000-0001-9649-7127} \and
Joyce El Haddad\inst{1}\orcidID{0000-0002-2709-2430}
\and
Daniela Grigori\inst{1}\orcidID{0000-0003-1741-8676}}
\authorrunning{W.Gherissi et al.}
%
\institute{Université Paris-Dauphine, Université PSL, CNRS, LAMSADE, 75016 PARIS, FRANCE\\ 
\email{\{wissam.gherissi,joyce.elhaddad,daniela.grigori\}@dauphine.psl.eu}\\
 \and {Telecom SudParis, Institut Polytechnique de Paris, PARIS, FRANCE \\ \email{mehdi.acheli@telecom-sudparis.eu}}}
\maketitle              

\begin{abstract}

Object-centric predictive process monitoring explores and utilizes object-centric event logs to enhance process predictions. The main challenge lies in extracting relevant information and building effective models.
In this paper, we propose an end-to-end model that predicts future process behavior, focusing on two tasks: next activity prediction and next event time. The proposed model employs a graph attention network to encode activities and their relationships, combined with an LSTM network to handle temporal dependencies. Evaluated on one real-life and three synthetic event logs, the model demonstrates competitive performance compared to state-of-the-art methods.


\end{abstract}

\section{Introduction}

Predictive Process Monitoring (PPM) is a key area of process mining focused on predicting the future evolution, potential deviations, and variations in processes.
The objective is to leverage advancements in machine learning and deep learning to develop models that can make accurate predictions based on event data, with a recent focus on object-centric event logs.
Traditional process mining models simplify event logs by associating each event with a single case, but this approach has limitations in real-world scenarios. Van der Aalst highlighted in \cite{convergence} 
shortcomings of this single case notion like convergence and divergence. Convergence occurs when the same event is linked to multiple cases, such as a shipping event tied to different items in the same package. Divergence arises when the same activity happens multiple times for a single case, such as picking and sorting multiple items for the same order. 


To overcome the aforementioned shortcomings, Object-Centric Event Logs (OCEL) were recently introduced in~\cite{ocelstd}. OCEL captures multiple perspectives of a process by relating events to various object types, offering richer insights compared to single-case event logs. This new object-centric data format has inspired advancements in traditional process mining techniques, including process discovery~\cite{discov}, its related metrics (precision and fitness)~\cite{precision},  conformance checking~\cite{constraints}, and predictive process monitoring. 
%
%
In PPM, recent works \cite{preserving}, \cite{defining}, \cite{GalantiLNM23} and \cite{wissam} have explored different ways to leverage OCEL’s object-centric information to improve predictions. One research direction, as in \cite{preserving}, focuses on creating process graphs that maintain the structural integrity of processes, avoiding the convergence and divergence problems where events relate to multiple or repeated cases. This method predicts the next event across the entire process without focusing on specific object types, which can lead to cumulative errors when predicting across different objects.
We argue for a different approach focusing predictions on a single object type at a time. This perspective-driven method, known as \emph{flattening} operation and defined in \cite{ocpm}, is useful when the analyst is primarily interested in one aspect of the process, such as the order in a retail system, rather than all related objects (e.g., individual items). 
For instance, if an analyst wants to predict the next event timestamp in an order process, this method directly targets order-related events without iterating through irrelevant item deliveries, thus reducing accumulation of deviations and prediction errors.
%
%
However, while focusing on a single object type, our perspective-driven approach  still benefits from considering interconnected object types. The dependencies between orders and items, for example, mean that item deliveries are influenced by order payments, and an order is only completed when all items are delivered. This relationship can be effectively modeled using an Object-Centric Directly-Follows Graph (OCDFG)~\cite{ocpm}, which captures both the activities and their connections to object types, allowing for more accurate predictions within the specific perspective.
%


In this paper, we propose a Graph Attention Network (GAT) to embed the nodes representing activities in the OCDFG, combined with a Long Short-Term Memory (LSTM) architecture for prediction. GATs excel at capturing the spatial relationships between nodes in a graph, enabling them to model interactions between activities across multiple object types. These graph embeddings serve as contextual information that enhances the LSTM model’s performance in two key prediction tasks: predicting the Next Activity and the Next Event Time. Both tasks are performed on the flattened event log of the selected object type.


The remainder of the paper is structured as follows. Section \ref{related} reviews related work on Object-Centric Process Mining (OCPM) and PPM. Section \ref{prelim} covers key concepts, including event logs, graph neural networks, and LSTM architectures. Section \ref{approach} details the proposed graph and LSTM model, from data preprocessing to predictions. Section \ref{exp} presents experimental results and discussions of findings, followed by conclusions in Section \ref{conclusion}.

\section{Related Work}\label{related}

Process mining, particularly predictive process monitoring, has evolved significantly with advancements in machine learning and deep learning. Various models have been adapted to predict future processes using techniques like sequence encoding with recurrent neural networks~\cite{rnn1, taxetal}, transformers \cite{transformer} and graph networks \cite{graphnn}. In recent years, graph structures and embeddings have gained prominence. 
%
For example, previous work~\cite{taco} combined graph encodings with Long Short-Term Memory (LSTM) architectures for process model graphs, while other research~\cite{ggnn} and \cite{pgtnet} focused on graph representations for event logs to predict remaining sequence time using gated graph neural networks or graph transformers. 
\cite{ggnn} used a gated graph neural network by transforming the prefix into a graph connecting events based on precedence (Forward, Backward, Repeat) whereas \cite{pgtnet} deploy a graph transformer architecture by using a GPS recipe \cite{gps}. 
However, these studies focused on traditional event logs, unlike the OCELs used in this paper.


The introduction of OCELs has opened up new possibilities for PPM, allowing for richer representations of processes by incorporating multiple object types within the same event log. One study \cite{preserving} analyzed OCELs using graph structures by defining process executions that grouped multiple objects and case notions. This method chose a leading object type and applied a distance threshold to limit object connections, which helped avoid an overwhelming number of events and related objects in the same execution. However, the prediction targets in such approaches are not specific to a particular object type, but rather predict the next activity independently of process perspectives.


Other works, such as those in~\cite{wissam} and~\cite{GalantiLNM23}, have aimed to improve prediction performance by adding object-centric features and context. They incorporated additional information, like the number of related objects for each event, to enrich the data and enhance prediction accuracy. In\cite{GalantiLNM23}, authors also constructed traces that included events related to specific objects and their interactions with other objects, to predict key performance indicators.
%
In contrast to these approaches \cite{wissam} and \cite{GalantiLNM23}, our paper introduces a method that incorporates object-centric relationships between activities in the form of graph embeddings, which are then passed through an LSTM architecture. Our approach allows for more accurate predictions by taking into account the relationships between object types, offering a distinct perspective compared to previous methods. Our evaluation study compares its performance against publicly available logs and implementations.


\section{Preliminaries}\label{prelim}


\subsection{Process Mining}
To introduce the definition of object-centric event logs, we
use some universes presented in \cite{ocpm}. Given $\mathbb{U}_{E}$ a universe of events, $\mathbb{U}_{act}$ a universe of activities, $\mathbb{U}_{o}$ a universe of object identifiers, $\mathbb{U}_{ot}$ a universe of object types, ans $\mathbb{U}_{time}$ a universe of timestamps, an object-centric event log can be defined as follow:



\begin{definition}[Object-Centric Event Log]\label{OCEL} An object-centric event log  $\mathcal{L}$ is defined as a tuple
$(E, O, OT, \pi_{act}, \pi_{time},\pi_{omap}, \pi_{otyp}, <)$ where,
\begin{itemize}
    \item $E \subseteq \mathbb{U}_E$ is the set of events, $O \subseteq \mathbb{U}_{o}$ is the set of object identifiers, and $OT \subseteq \mathbb{U}_{ot}$ is the set of object types. 
    \item $\pi_{act} : E \to \mathbb{U}_{act}$ is the function associating each event with its activity.
    \item $\pi_{time} : E \to \mathbb{U}_{time}$ is the function associating each event with a timestamp.
    \item $\pi_{omap} : E \to \mathcal{P}(O)$ is the function associating an event identifier to a set of related object identifiers.
    \item $\pi_{otyp} \in O \to OT$ is the function assigning precisely one object type for each object identifier.
    \item $<$ is a partial order based on the timestamps of events such that
    $$e_1 < e_2 \iff \pi_{time}(e_1) < \pi_{time}(e_2) $$
    \end{itemize}
\end{definition}

\begin{definition} [Flattened Log] \label{flatten}
Let $\mathcal{L}$ be an object-centric event log, and $ot \in O T$ be an object type. We define the flattened log of $\mathcal{L}$  over $ot$ as  $\mathcal{FL} (\mathcal{L},ot)=\left(E^{ot}, \pi_{act}^{ot}, \pi_{time }^{ot}, \pi_{case}^{ot}, \leq^{ot}\right)$ where:
\begin{itemize}
    \item $E^{o t}=\left\{e \in E \mid \exists{o \in O} \pi_{o t y p}(o)=\right.$ ot $\left.\wedge o \in \pi_{omap}(e)\right\}$
    \item $\pi_{a c t}^{o t}=\left.\pi_{a c t}\right|_{E^{o t}}$
    \item $\pi_{time}^{o t}=\left.\pi_{t i m e}\right|_{E^{o t}}$
    \item For $e \in E^{o t}, \pi_{case}^{o t}(e)=\left\{o \in \pi_{omap}(e) \mid \pi_{o t y p}(o)=o t\right\}$
    \item $<^{o t}=\left\{e_1<e_2 \mid \exists{o \in O} \pi_{o t y p}(o)=\right.$ ot $ \wedge o \in$ $\left.\pi_{omap}\left(e_1\right) \cap \pi_{omap} \left(e_2\right)\right\}$
\end{itemize}
\end{definition}

The flattening of OCEL with an object-type will allow us to define prediction tasks specific to an object-based perspective.

\begin{definition}[Directly-Follows Graph]
A directly-follows graph $\mathcal{DFG}$ is a tuple $\left(A, F, \pi_{freqe}\right)$ where:
\begin{itemize}
    \item $A \subseteq U_{a c t}$ is a set of activities,
    \item $F \subseteq A \times A$ is the set of edges,
    \item $\pi_{freqe}: F \nrightarrow \mathbb{N}$ is a frequency measure on the edges.
\end{itemize}
\end{definition}

Note that in this definition, we use the version of DFG considering only the frequency measures on edges.

\begin{definition}[Discovery of DFG] \label{discov dfg}
Let $\mathcal{FL} =\left(E^{ot}, \pi_{act}^{ot}, \pi_{time }^{ot}, \pi_{case}^{ot}, \leq^{ot}\right)$ be a flattened event log,  its associated directly-follows graph $\mathcal{DFG(FL)}=\left(A, F, \pi_{f r e q e}\right)$ is constructed following the discovery operation where :
\begin{itemize}
    \item $A=\pi_{act}^{ot}(\mathcal{FL})$,
    \item $F= \{\cup_{c \in \pi_{case}^{ot}(\mathcal{FL}), \mathcal{FL}(c)=\langle a_1, \ldots, a_n\rangle}\left\{\left(a_i, a_{i+1}\right) \mid 1 \leq i<n\right\}$
    \item For $(a, b) \in F \cap(A \times A)$,
    $$\begin{aligned}
    & \pi_{freqe}(a, b)=\sum_{\substack{c \in \pi_{case}^{ot}(\mathcal{FL}), \\ \mathcal{FL}(c)=\left\langle a_1 \ldots, a_n\right\rangle}} & \left|\left\{i \in \mathbb{N} \mid 1 \leq i<n \wedge a_i=a \wedge a_{i+1}=b\right\}\right|
    \end{aligned}$$
\end{itemize}
\end{definition}




\begin{definition}[Object-Centric Directly-Follows Graph] \label{def_ocdfg}
An object-centric directly-follows graph $\mathcal{OCDFG}$ is a tuple $\left(A, OT, F, \pi_{\text {ofreqe}}\right)$ where:
\begin{itemize}
    \item $A$ is a set of activities,
    \item $OT$ is a set of object types,
    \item $F \subseteq  A \times A $ is a set of  edges,
    \item $\pi_{\text {ofreqe }}: F \times OT \nrightarrow \mathbb{R}^{+} $ assigns for each edge and object type a frequency measure.
\end{itemize}
\end{definition}

\begin{definition}[Discovery of OCDFG]\label{discov ocdfg}
Given an OCEL $\mathcal{L}$, and a set of DFGs of flattened logs
$\{G_i \ | \ G_i=\mathcal{DFG}(\mathcal{FL}(\mathcal{L}, ot_i))=(A_i, F_i, \pi_i)$, for $i = 1 \cdots |OT|\}$, the discovery operation of $\mathcal{OCDFG}(\mathcal{L})$
=$\left(A, O T, F, \pi_{\text {ofreqe}}\right)$
where for given $n1$, $n2$ $\in A$ we have  $(n1, n2) \in F \Leftrightarrow \exists o_i \in OT$ such that  $(n1, n2) \in F_i$, and where 
edge frequencies for each object type are computed as follows : 
$\forall o_i \in OT$, if 
$(n1, n2) \in F_i$
then $\pi_{\text {ofreqe }}(n1,n2, o_i) = \pi_i(n1,n2)$;
else $\pi_{\text {ofreqe }}(n1,n2, o_i) = 0$.
The discovery is summed up in the following steps:
\begin{itemize}
    \item Flatten OCEL for each object type to allow an object type-based prediction, 
    \item Perform discovery of DFG  for each flattened event log (Def.\ref{discov dfg}),
    \item Merge edges of the resulting DFGs between each pair of nodes. Each edge will have a frequency associated to each object type, equal to the frequency of the edge of the corresponding DFG of the flattened log or 0 if no edge exist. The resulting OCDFG aggregates dependencies between activities for different object types. 
\end{itemize}

\end{definition}

\subsection{Neural networks}
\subsubsection{Graph Attention Networks (GAT)}\cite{graphattentionnetworks} are a type of Graph Neural Network (GNN) that incorporate attention mechanisms to process graph-structured data.
Unlike traditional GNNs, which apply fixed or uniform weights to aggregate information from neighboring nodes, GAT dynamically learns to weigh the importance of each neighboring node during message-passing.
This is useful when nodes vary in their relevance to the prediction task.
%
In PPM, processes can be represented as graphs, where nodes correspond to activities and edges reflect inter-relationships, such as causal dependencies, or directly-follows relationships in our case. GAT  focuses on the most informative event relationships, filtering out less relevant connections, and assigns different weights to neighboring nodes, based on their influence on the outcomes. 
By capturing these complex dependencies, GAT improves the accuracy and interpretability of prediction models. Its ability to highlight important interactions within event logs makes GAT an effective tool for understanding and predicting process behaviors.

\subsubsection{Long Short Term Memory (LSTM)} networks are a type of Recurrent Neural Network (RNN) designed to overcome the limitations of traditional RNNs in modeling long-term dependencies. LSTMs use a gating mechanism to control the flow of information, allowing them to retain important information over long time horizons while mitigating the vanishing gradient problem. 
LSTMs use a gating mechanism to manage information flow, retaining essential details over extended periods and solving the vanishing gradient problem. This makes LSTMs optimal for handling sequential data, such as event sequences in PPM, where past events influence future outcomes.
By maintaining a memory of past events, LSTMs are able to predict future events or outcomes more accurately, making them an ideal candidate for our use-case.

To conclude, the combination of GAT and LSTM leverages the strengths of both architectures. GAT focuses on the relational aspect of activities in an OCDFG by learning the dependencies between them, while LSTM models the temporal dynamics of the process sequence. By combining these two approaches, the model can simultaneously learn both the structural (graph-based) and temporal (sequence-based) characteristics of event logs.
 


\section{Our approach}\label{approach}

Our goal is to develop an end-to-end model combining a Graph Attention Network (GAT) with LSTM layers to predict future behavior in a flattened event log for a specific object type. In real-life business scenarios, tracking and predicting each object's lifecycle is crucial. To achieve this, we apply a flattening operation, focusing on individual object types while still utilizing inter-object contextual information from the OCEL via the OCDFG. Our approach consists of two main steps: preprocessing and prediction, each detailed further in the following sections.

\subsection{Step 1 : Preprocessing}

Below, we outline the key operations involved in the preprocessing step, which prepares the model’s inputs. Starting with the OCEL as the initial input, we perform a flattening operation (see Def.\ref{flatten}) on each object type in the OCEL to generate single object-type event logs. For the selected object type and its corresponding flattened log, we extract the temporal features and prediction targets, as further detailed and shown in Fig.\ref{fig:general dataflow}(a). Simultaneously, we use the collection of flattened event logs to discover the OCDFG (see Def.\ref{discov dfg}), as illustrated and explained in Fig.\ref{fig:general dataflow}.

\subsubsection{Temporal features and prediction targets.}\label{TMP}
We begin by flattening the OCEL ($\mathcal{L}$) on different object types. For a given flattened event log 
($\mathcal{FL}(OT_i)$), we extract each object sequence and generate multiple prefixes of varying lengths. For example, for object type \emph{item}, if the object is $i2$, we extract a prefix of length $3$: \texttt{Place Order(PO)}, \texttt{Pick Item (PI)}, \texttt{Send Delivery(SD)}.\\
For each event in the prefix, we compute the following temporal features: \emph{time since the last event}, \emph{time since the start of the prefix}, \emph{time since midnight} of the same day of the event and \emph{weekday of the event}.\\
As prediction targets, we retain information on the next activity following the prefix (NA) and the time until the execution of that next activity (NE).\\
Finally, we apply average scaling to all temporal features and time targets to maintain computational stability while preserving the temporal distribution..

\subsubsection{Object-centric directly-follows graph.} \label{transformation}
\begin{figure*}[t]
    \centering
    \centerline{\includegraphics[scale=0.5]{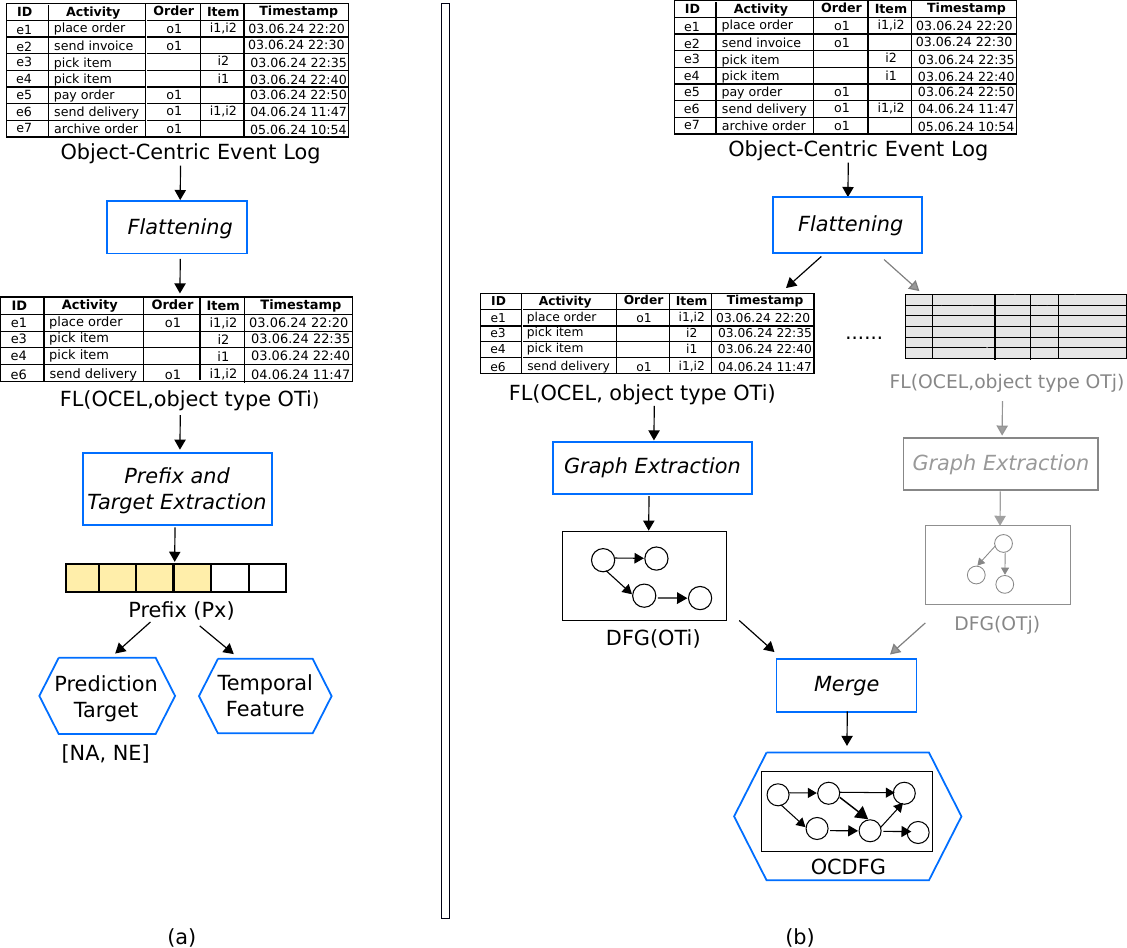}}
    \caption{(a) Temporal features and prediction targets  preprocessing, and (b) OCDFG construction }
    \label{fig:general dataflow}
\end{figure*}

From each flattened log, we discover a DFG (see Def. \ref{discov dfg}).  We then merge these individual DFGs to construct an Object-Centric Directly-Follows Graph (OCDFG) (see Def.\ref{discov ocdfg}). The resulting OCDFG captures the connections between activities across all object types and flattened event logs, as illustrated in  Fig.\ref{fig:general dataflow}(b).
Specifically, each edge between consecutive activities represents all object types and their corresponding frequencies. This resulting graph serves as the input for the GAT in the model.

\begin{figure}[t]
    \centering
    \centerline{\includegraphics[scale=0.5]{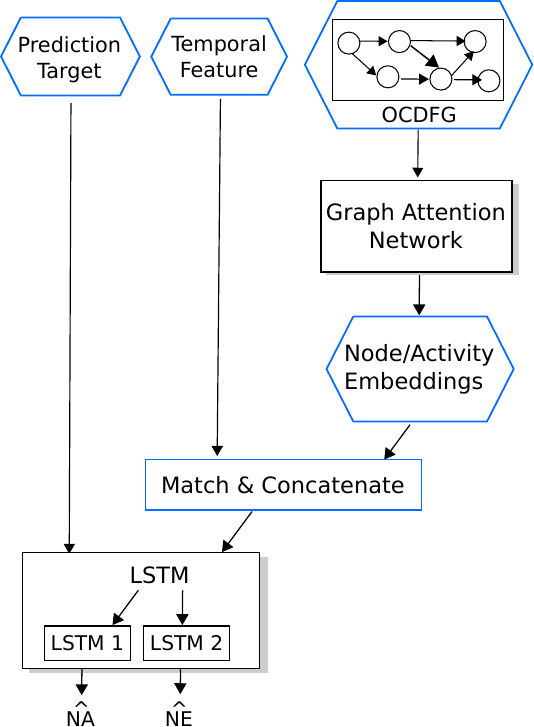}}
    \caption{Application of the approach on object type OT$_i$}
    \label{fig:object type specific}
\end{figure}

\subsection{Step 2 : Prediction model}
In this step, first, we feed OCDFG as input for the GAT to generate embeddings for the nodes representing the activities. Next, we perform a \emph{Match and Concatenate} operation combining the temporal features of the events with their corresponding graph embeddings. These combined features are then fed into the LSTM network to make predictions for the next activity ($\widehat{NA}$) and the time until the next event ($\widehat{NE}$).
%
This approach ensures smooth gradient flow through the pipeline, allowing both the GAT and LSTM networks to be updated. The overall process is illustrated in Fig.\ref{fig:object type specific}.\\
The neural network model consists of two components: a Graph Attention Network (GAT) and an LSTM architecture. The GAT generates node embeddings from the OCDFG, while the LSTM captures the sequential information from prefixes to predict future behavior. This structure is depicted in Fig.~\ref{fig:object type specific}. Further details of the model are provided below.
\begin{itemize}
    \item \textbf{Graph Attention Network} consist of three Graph Attention Convolution layers, with variable hidden and embedding output size depending on the event log used. The OCDFG is
    passed through this network to generate node embeddings, which encode each activity. 
    For a given sequence of events, each event has an associated activity. We perform a differential matching operation, aligning each activity in the sequence with its corresponding embedding, and then concatenating the selected embedding vector with the other event features.   
    For instance, consider an event \texttt{e1} with the activity \texttt{Place\_Order}. In the OCDFG embeddings, the node \texttt{PO}, for actvitiy \texttt{Place\_Order}, is represented by the vector \texttt{V1}. We match this node with the activity \texttt{Place\_Order} in the event and concatenate the embedding vector \texttt{V1} with the temporal features of \texttt{e1}, resulting in the final feature vector.
    
    \item \textbf{LSTM network} consists of a shared LSTM layer with hidden and output size 100, followed by two other specialized LSTM layers, each dedicated to a specific prediction task. Both layers have a hidden size of 100. 
    The specialized layer for Next Activity prediction outputs the probabilities for each possible activity, while the Next Event Time prediction layer has an output size of 1 for the regression task. The input to the LSTM network is a sequence prefix, where each event is represented by its activity embedding and temporal features.
\end{itemize}

\section{Experimental results}\label{exp}
In this section, we evaluate our model (\textbf{LSTM+GAT}) and compare it against two publicly available prediction models: one-hot encoding+LSTM  of \cite{wissam} (\textbf{LSTM}) and ProcessTransformer of \cite{transformer} (\textbf{PT}). The evaluation is performed using four OCELs, including a real-life event log for loan applications from the BPIC17 dataset\footnote{\label{footnote}\url{https://data.4tu.nl/articles/dataset/BPI_Challenge_2017/12696884}} and three synthetic logs simulating an order management process, a logistics process and a Purchase-to-Pay process\footnote{Made available by \url{https://www.ocel-standard.org/event-logs}}.

The experiments were conducted on a server equipped with AMD EPYC 7702 64-Core Server Processor and GeForce RTX 2080 Ti.
The code and results are reproducible and can be accessed here\footnote{\url{https://github.com/wissam-gherissi/GATLSTM}}.

\subsection{Event logs}

\textit{Order Management.} describes an order management process. It contains 22367 events, 11522 objects and 5 object types. We focus only on three specific object types that are essential to the process which are: \textbf{item}, \textbf{order} and \textbf{package}. Hence, we flatten the OCEL input  based only on these object types.

\medskip
\noindent \textit{{Logistics.}} describes the logistics process of a company that sells goods overseas. It contains 35761 events, 14013 object and 7 object types. In this log, we focus on object types with processes containing at least three different types of activities, otherwise, the next activity prediction can be considered meaningless if there is one possible future activity. We consider only the predictions for the following object types : \textbf{container}, \textbf{transport doc}ument and \textbf{vehicle}.

\medskip
\noindent \textit{{Purchase-to-Pay.}} describes the Procure-to-Pay process within an organization following the creation of a purchase until the execution of the payment. It contains 14671 events, 9543 objects and 7 object types. Similarly, we consider only processes with more than two different types of activities. We consider only the following object types: \textbf{goods} receipt, \textbf{invoice} receipt, \textbf{purch}ase \textbf{order}, \textbf{purch}ase requisition, and \textbf{quotation}. 

\medskip
\noindent \textit{{BPIC17.}} is a real-life event log describing a loan application process in a financial institution following the process of applications and the offers provided by the institution. It contains 507553 events, 67498 objects and 2 object types: \textbf{application} and \textbf{offer}.

\subsection{Results}
%


\paragraph{Next Activity Prediction.}

\begin{table}[!t]
    \centering
\begin{tabular}{|c|c|c|c|c|c|c|c|c|c|c|}
\hline 
\multirow{2}{*}{Event logs} & \multirow{2}{*}{ Object Type } & \multicolumn{3}{|c|}{ Model } \\
\cline{3-5} & & LSTM & PT & \textbf{LSTM+GAT}  \\
\hline 
\multirow{3}{7em}{Order Management} & Order & 58.7(54.4) &\textbf{72.2(65.4)} & 59.6(52.2)\\
\cline{2-5} & Item & 73.1(66.5) &76.1(64.4) &  \textbf{77.4(73.6)} \\
\cline{2-5} & Package & 90.4(87.6) &82.3(71.1) & \textbf{90.5(86.3)}\\
\hline \multirow{2}{5em}{BPIC17} & Application & 87.2(85.4) &82.3(71.1) & \textbf{88.0(86.1)} \\
\cline{2-5} & Offer & 81.7(74.5) & 79.4(73.0) & \textbf{81.7(74.9)} \\
\hline \multirow{3}{5em}{Container Logistics} & Container & 95.6(95.4) & 97.3(96.0) & \textbf{97.3(97.2)}\\
\cline{2-5} & Transport Doc & 97.4(96.2) & \textbf{98.3(97.1)}  & 97.4(96.2)  \\
\cline{2-5} & Vehicle & 87.9(84.6) & 86.4(86.5) & \textbf{88.0(84.8)}  \\
\hline \multirow{6}{5em}{P2P} & Goods & 76.4(71.5) & 74.2(65.0) & \textbf{76.4(71.6)}\\
\cline{2-5} & Invoice & 81.0(80.1)&  \textbf{85.8(78.2)} &80.7(79.5)  \\
\cline{2-5} & Purch Order & 83.0(82.6) &  70.4(67.8) &\textbf{83.2(82.3)}  \\
\cline{2-5} & Purch Requisition & 92.0(88.3) & \textbf{93.4(85.7)} &92.0(88.3)  \\
\cline{2-5} & Quotation & 81.1(81.1) & \textbf{87.9(82.3)} &82.4(82.2)  \\
\hline
\end{tabular}
    \caption{Accuracy (F1 score) expressed in percentages for Next Activity prediction task }
    \label{tab:next_act}
\end{table}


In Tab.\ref{tab:next_act}, we see that our model consistently outperforms the basic LSTM model across all event logs and object types. This performance improvement is closely tied to the difference in activity encoding methods. The integration of the GAT model, which incorporates object-centric information, provides deeper insights that enhance the accuracy of activity predictions.
%
%
Compared to ProcessTransformer, our model and the basic LSTM perform better on three out of the four event logs (excluding Order and Transport Doc). This highlights the robustness of the LSTM architecture for Next Activity prediction. However, for the specific object types where ProcessTransformer outperforms our model or the LSTM model, we observe that LSTM-based models tend to struggle. This underperformance suggests that LSTMs can suffer from issues like loops, overfitting, or converging to local optima, which limit their ability to predict the next activity effectively in certain cases.

\paragraph{Next Event Time Prediction.}

\begin{table}[!t]
    \centering
\begin{tabular}{|c|c|c|c|c|c|c|c|c|c|c|}
\hline 
\multirow{2}{*}{Event logs} & \multirow{2}{*}{ Object Type } & \multicolumn{3}{|c|}{ Model } \\
\cline{3-5} & & LSTM & PT & \textbf{LSTM+GAT}  \\
\hline 
\multirow{3}{8em}{Order Management} & Order&0.97 &1.60 & \textbf{0.95}\\
\cline{2-5} & Item &\textbf{0.94}  &1.63 & 1.29  \\
\cline{2-5} & Package &0.32 &\textbf{0.26} & 0.37 \\
\hline \multirow{2}{6em}{BPIC17} & Application &0.42 &0.41  & \textbf{0.41} \\
\cline{2-5} & Offer & 1.28& 2.55 & \textbf{1.28}\\
\hline \multirow{3}{5em}{Container Logistics} & Container & 0.69 & 0.74 & \textbf{0.69}\\
\cline{2-5} & Transport Doc & \textbf{0.59} & 2.7 & 0.60  \\
\cline{2-5} & Vehicle & 0.64 & \textbf{0.61} & 0.67  \\
\hline \multirow{6}{5em}{P2P} & Goods & 0.81 & 0.92 & \textbf{0.80}\\
\cline{2-5} & Invoice & 0.44 & 0.91 & \textbf{0.43}  \\
\cline{2-5} & Purch Order & 1.66 & \textbf{1.50} & 1.67  \\
\cline{2-5} & Purch Requisition & \textbf{1.03} & 1.40 &1.07  \\
\cline{2-5} & Quotation & \textbf{0.55} & 1.00 & 0.58  \\
\hline

\end{tabular}
    \caption{Mean Absolute Error (MAE) expressed in days for Next Event Time prediction task}
    \label{tab:next_event}
\end{table}

In Tab.\ref{tab:next_event}, we observe a slight improvement in our model’s performance for predicting the next event time compared to the LSTM model. The use of graph embeddings adds valuable information about activities and their interrelations for each object type. However, this improvement is not substantial or consistent across all tested event logs. This suggests that while the directly-follows relationship between activities helps, the primary influence still comes from the temporal features extracted from the prefix. Our model outperforms the others on at least half of the event logs and object types. The LSTM model follows as the second best, with ProcessTransformer performing best only on three object types.

\subsection{Discussion}\label{discuss}


This section discusses the proposed model and draws conclusions from the experimental results. The end-to-end nature of the model allows the GAT to update activity embeddings based on the prediction loss, leading to better representation of activities and their relationships using the OCDFG for capturing contextual information.
%
%
The analysis of results in Tab.\ref{tab:next_act} shows significant improvement and superior performance of the LSTM+GAT model compared to both LSTM and PT models. However, there is room for improvement in addressing issues like loops and overfitting, particularly with the \textbf{Order} object type.
%
%
Additionally, Tab.\ref{tab:next_event} 
highlights the robust performance of the LSTM architecture for sequential prediction, with LSTM-based models outperforming PT in the Next Event Time task. However, the slight improvement from the baseline LSTM to LSTM+GAT suggests that incorporating additional temporal features may enhance model performance further.

Overall, the experiments demonstrate promising results in two key aspects: the effectiveness of LSTM for predictions and the use of graph embeddings to represent activities via the OCDFG. ProcessTransformer was outperformed by LSTM models in both Next Activity and Next Event Time prediction tasks, except for a few object types in the latter task.
The superior performance of LSTM+GAT over LSTM, particularly in Next Activity prediction, confirms the advantage of using object-centric graph embeddings instead of one-hot encoding. To improve Next Event Time prediction, incorporating temporal information, such as average running times and waiting times in the OCDFG, is suggested.

\section{Conclusion and future work} \label{conclusion}
In this paper, we proposed a solution for predictive monitoring using object-centric logs. By leveraging a graph model to embed activities based on an OCDFG, we introduced an end-to-end architecture that outperformed baseline models, including a simple LSTM and ProcessTransformer. Our approach focused on each object type by flattening the OCEL while utilizing the object-centric information in the OCDFG. We discussed the key findings and model performance compared to state-of-the-art models. Future work will explore more real-life OCELs, unsupervised embeddings like Node2Vec, and time-related features in the OCDFG. Additionally, we will investigate advanced graph neural networks, such as Dynamic Graph Neural Networks, to capture evolving patterns and enhance Next Event Time prediction.


\bibliographystyle{splncs04} 
\bibliography{bib.bib}

\end{document}